\documentclass[10pt,twocolumn,letterpaper]{article}
\usepackage{iccv}
\usepackage{multirow}
\usepackage{booktabs} 
\usepackage{placeins}
\usepackage{multirow} 
\usepackage{adjustbox} 
\usepackage{graphicx}
\usepackage{pgfplots}
\usepackage{caption}
\usepackage{CJKutf8}
\usepackage{algorithm}
\usepackage{algpseudocode}
\usepackage{bm}
\pgfplotsset{compat=1.18}
\definecolor{iccvblue}{rgb}{0.21,0.49,0.74}
\usepackage[pagebackref,breaklinks,colorlinks,allcolors=iccvblue]{hyperref}
\usepackage[accsupp]{axessibility}

\title{
FakeRadar: Probing Forgery Outliers to Detect Unknown Deepfake Videos
}
\author{Zhaolun Li$^{1}${\hspace{2.5em}}Jichang Li$^{2*}${\hspace{2.5em}}Yinqi Cai$^{3}${\hspace{2.5em}}Junye Chen$^{3}${\hspace{2.5em}}\\Xiaonan Luo$^{1}${\hspace{2.5em}}Guanbin Li$^{3,2,5}${\hspace{2.5em}}Rushi Lan$^{1,4}\thanks{Corresponding Authors.}$ \vspace{0mm}\\
$^1$Guilin University of Electronic Technology\quad $^2$Pengcheng Laboratory\quad $^3$Sun Yat-sen University\quad \\
$^4$Guangxi Key Laboratory of Image and Graphic Intelligent Processing \vspace{0mm} \\
$^5$Guangdong Key Laboratory of Big Data Analysis and Processing \vspace{0mm} \\
{\tt\small zhaolunli@mails.guet.edu.cn, li.jichang@pcl.ac.cn, rslan2016@163.com}
 \vspace{0mm}
}

\begin{document}
\maketitle

\begin{abstract}
In this paper, we propose FakeRadar, a novel deepfake video detection framework designed to address the challenges of cross-domain generalization in real-world scenarios. Existing detection methods typically rely on manipulation-specific cues, performing well on known forgery types but exhibiting severe limitations against emerging manipulation techniques. This poor generalization stems from their inability to adapt effectively to unseen forgery patterns. To overcome this, we leverage large-scale pretrained models (e.g. CLIP) to proactively probe the feature space, explicitly highlighting distributional gaps between real videos, known forgeries, and unseen manipulations. Specifically, FakeRadar introduces Forgery Outlier Probing, which employs dynamic subcluster modeling and cluster-conditional outlier generation to synthesize outlier samples near boundaries of estimated subclusters, simulating novel forgery artifacts beyond known manipulation types. Additionally, we design Outlier-Guided Tri-Training, which optimizes the detector to distinguish real, fake, and outlier samples using proposed outlier-driven contrastive learning and outlier-conditioned cross-entropy losses. Experiments show that FakeRadar outperforms existing methods across various benchmark datasets for deepfake video detection, particularly in cross-domain evaluations, by handling the variety of emerging manipulation techniques.
\end{abstract}

\section{Introduction}
\label{sec:intro}

\begin{figure}[!t]
\centering 
\includegraphics[width=1.0\columnwidth]{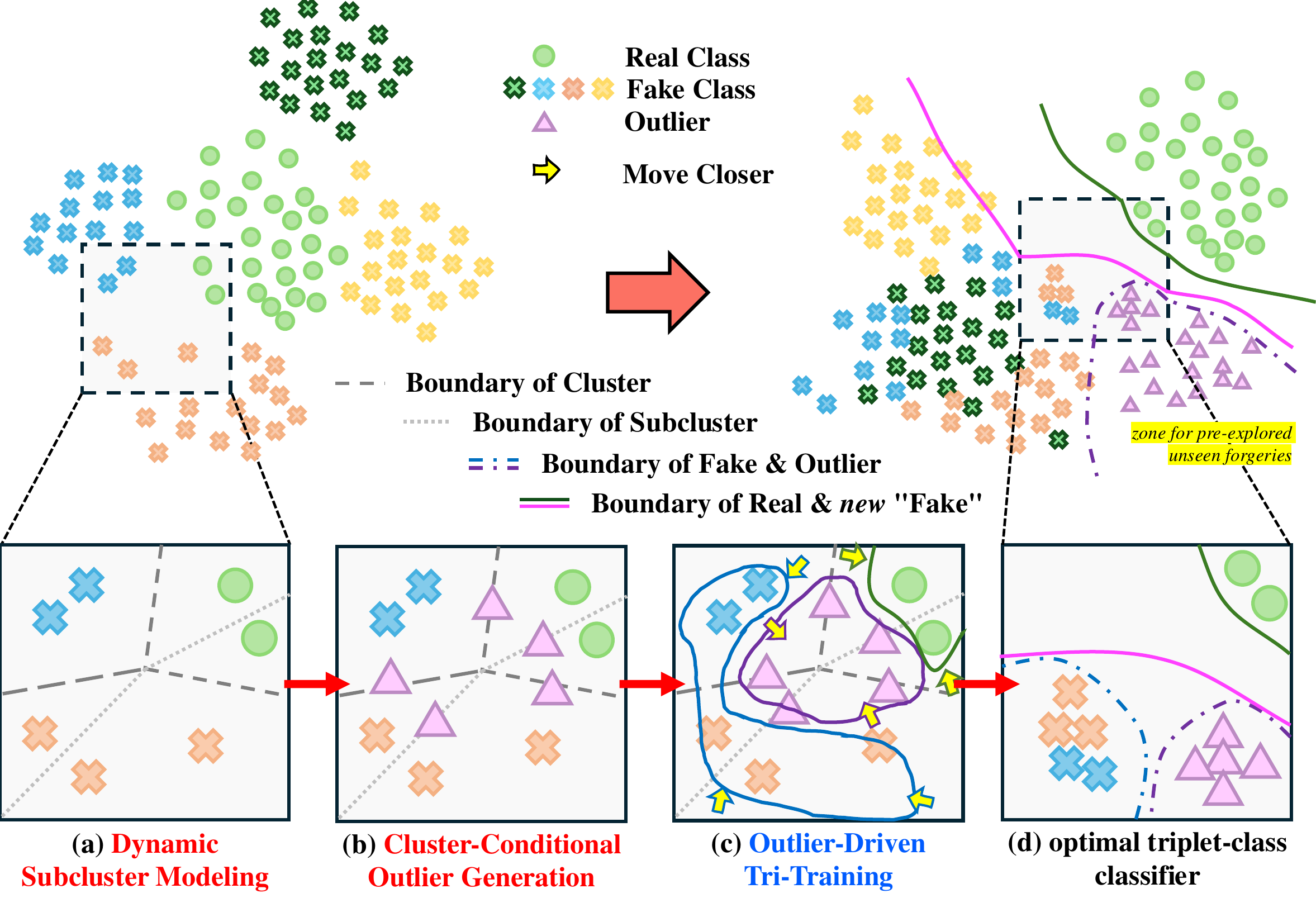} 
\vspace{-15pt}
\caption{
Conceptual pipeline of FakeRadar. 
(a)-(b) Forgery Outlier Probing first performs dynamic subcluster modeling on these features to partition finer-grained subclusters, then generates cluster-conditional outliers for each estimated subcluster to simulate unseen forgeries, emulating unknown manipulations. (c)-(d) The model is optimized using outlier-driven tri-training, allowing the detector to establish optimal boundaries for the real, fake, and outlier classes, with the latter two merged into a new ``Fake'' class during testing.
} 
\label{fig:intro} 
\vspace{-10pt}
\end{figure}

Recently, generative techniques~\cite{generate_SketchEdit,generate_DINO-ViT,generate_UV-IDM} for creating deepfake videos have advanced rapidly, producing highly realistic facial forgeries that challenge human perceptual limits~\cite{challenge_human_perceptualimits_1,challenge_human_perceptualimits_2}. However, the malicious use of such technology, such as fabricated news and identity fraud, poses a significant threat to digital media credibility~\cite{IDGuard,Deepfakes_crime}.
Earlier deepfake detection methods primarily rely on the passive learning of specific forgery artifacts, such as boundary inconsistencies~\cite{LSDA,SeeABLE}, blinking anomalies~\cite{eyes-blinking}, and texture irregularities~\cite{Gu2021ExploitingFF,LAA-NET}. While these approaches perform well against conventional deepfake manipulations, including Face2Face~\cite{F2F} and NeuralTextures~\cite{NeuralTextures}, they exhibit critical vulnerabilities when confronted with emerging generative paradigms, such as diffusion models~\cite{yang2023diffusion}. For instance, boundary-based detection methods~\cite{X-Ray,SBI} fail to identify faces generated by advanced GANs (e.g., StyleGAN~\cite{Stylegan}). Similarly, blink analysis-based techniques~\cite{eyes-blinking} are rendered ineffective when processing temporally smoothed fake videos~\cite{temporally_smoothed_fake_videos}. These limitations stem from a fundamental shortcoming: existing methods learn a constrained decision boundary in the feature space, where known forgery traces occupy only a localized region, while novel manipulation techniques introduce artifacts in arbitrary locations. 
As a result, they struggle to generalize to new manipulation techniques that introduce artifacts in previously unseen ways, leading to significant performance degradation in cross-domain scenarios~\cite{DFDC,DFDCP}.

To better understand its underlying cause, \cite{universaldetect,RECCE} have analyzed the feature distributions of real and synthetic data. These studies reveal a fundamental embedding-space disparity: real data exhibit compact, dense distributions, whereas deepfake forgeries form scattered, sparse clusters due to their generation with diverse manipulation types, as demonstrated in Figure~\ref{fig:intro}. This difference in distribution poses a significant challenge for existing detection methods, which are typically optimized to classify previously observed forgeries. Since deepfake forgeries do not adhere to a fixed pattern but rather introduce an infinite variety of modifications, classifiers trained on known forgery patterns struggle to recognize novel deepfakes that deviate from past observations. Therefore, a robust detection approach must move beyond merely learning pre-defined forgery patterns and instead develop the capability to identify anomalies that fall outside of previously known distributions.
Given the limitations of existing methods in handling unseen forgery types, we propose a paradigm shift in deepfake detection. Instead of passively fitting to known forgery patterns, our approach proactively explores potential anomaly regions in the feature space. Inspired by radar systems that scan for unknown targets through frequency-spectrum probing, our method enhances detection robustness against novel deepfake manipulations.

In this work, we propose a new deepfake video detection framework, called \textit{FakeRadar}, which leverages the deep feature priors of large-scale pre-trained models (i.e. CLIP~\cite{Clip}) to proactively ``probe'' unknown forgeries and enhance cross-domain generalization. Unlike conventional detectors~\cite{RECCE,StyleFlow,AltFreezing} that perform only binary classification, FakeRadar explicitly introduces ``forgery outliers'' during training to simulate potential unknown forgeries and thus ``pre-explore and cover'' the latent space. As shown in Figure~\ref{fig:intro} (a) and (b), we first employ Forgery Outlier Probing, which conducts dynamic subcluster modeling on the features of real and known deepfake videos. By approximating the embedding space via fine-grained Gaussian mixture distributions, this algorithm automatically partitions training samples into distinct subclusters. Next, through cluster-conditional outlier generation, Forgery Outlier Probing synthesizes outlier samples near the boundaries of these estimated subclusters with a certain ``novel offset'', thereby covering a broader range of potential forgery regions. 

Additionally, FakeRadar adopts a model optimization strategy called Outlier-Guided Tri-Training, enabling the model to discriminate among ``Real'', ``Fake'', and ``Outlier'' samples, with its details as illustrated in Figure~\ref{fig:intro} (c) and (d). Specifically, Outlier-Guided Tri-Training proposes an outlier-driven contrastive loss and an outlier-conditioned cross-entropy loss for end-to-end model optimization. The former loss enhances the distance between real, fake, and outlier features via contrastive learning, emphasizing their differences, while the latter one ensures that the model has clear decision boundaries for the three categories, especially ensuring that outliers are not incorrectly classified as real. 
In this triplet-class (Real/Fake/Outlier) setting, the model can independently label unknown forgeries during training. During inference, both ``Fake'' and ``Outlier'' are classified as forgeries, empowering the detector to adapt effectively to unknown deepfakes.
We summarize FakeRadar's principal contributions as follows, with its conceptual pipeline illustrated in Figure~\ref{fig:intro}:
\begin{itemize}
    \item 
    We propose \textit{FakeRadar}, a novel framework for deepfake video detection, designed to address the critical challenge of cross-domain generalization in real-world scenarios.
    \item We introduce Forgery Outlier Probing for synthesizing outlier samples to cover a broader range of potential forger patterns, expanding the detector's understanding of forgery distribution.
    \item We design Outlier-Guided Tri-Training for model optimization, which enables end-to-end training for distinguishing real, fake, and outlier samples, improving detection of unknown forgeries.
    \item Numerous experiments have revealed that FakeRadar outperforms current leading algorithms across multiple deepfake benchmarks, particularly excelling in cross-domain evaluations where existing methods struggle.
\end{itemize}

\section{Related work}
\label{sec:Related work}

\begin{figure*}[h!]
\centering
\includegraphics[width=1.0\textwidth]{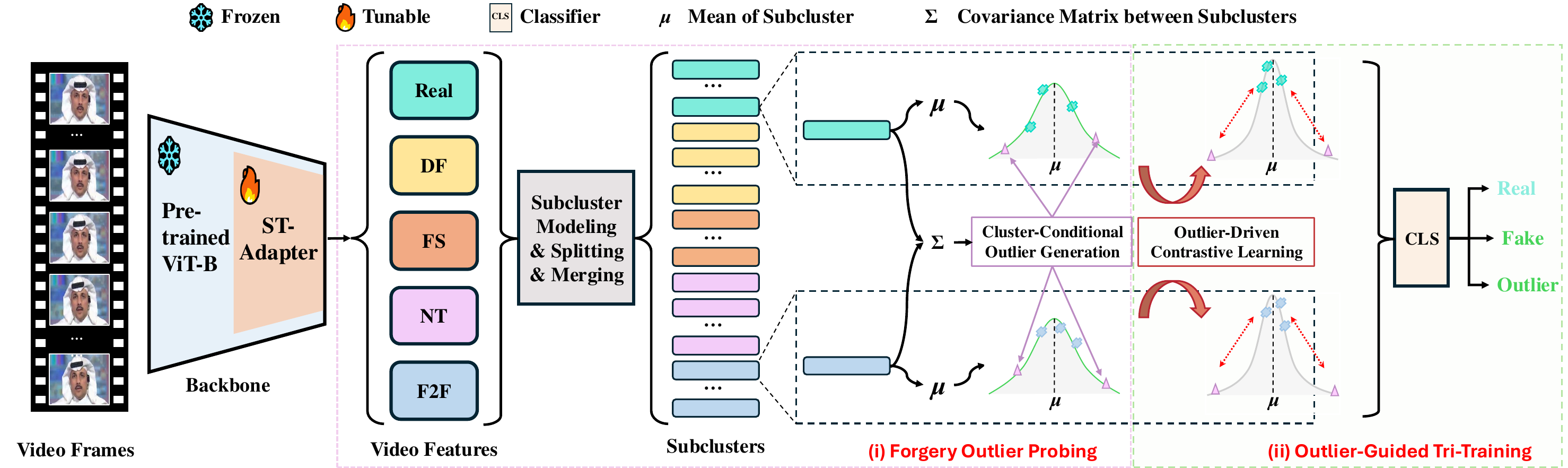} 
\vspace{-15pt}
\caption{
An overview of FakeRadar for deepfake video detection. 
FakeRadar consists of two key components: {Forgery Outlier Probing} and {Outlier-Guided Tri-Training}. (i) {Forgery Outlier Probing} performs dynamic subcluster modeling to model the feature-space distributions of real and fake samples using GMMs, followed by dynamic subcluster splitting and merging to produce finer-grained subcluster partitions. Afterwards, FakeRadar conducts cluster-conditioned outlier generation to simulate unseen forgeries for novel manipulation techniques. (ii) {Outlier-Guided Tri-Training} jointly optimizes the CLIP model’s ViT-B/16 backbone, inserted with ST-Adapter~\cite{st-adapter} for parameter-efficient fine tuning, and a triplet-class classifier to distinguish between ``Real'', ``Fake'', and ``Outlier'' classes, with ``Fake'' and ``Outlier'' merged into a single new ``Fake'' class during evaluation. The training set includes real videos, as well as deepfake videos generated using four different manipulations, i.e. ``DF''~\cite{DFD},  ``FS''~\cite{faceswap}, ``NT''~\cite{NeuralTextures} and ``F2F''~\cite{F2F}.
} 
\label{fig:framework} 
\vspace{-10pt}
\end{figure*}

\paragraph{Deepfake Detection.}
In response to the malicious use of deepfake techniques, numerous approaches have been explored to address this issue. Generally, deepfake detection is formulated as a binary classification problem (real or fake), leading researchers to develop a variety of end-to-end detectors~\cite{DCL,LSDA} that directly distinguish reals from manipulated content. Existing detection methods can be broadly categorized into two groups: image-level and video-level detection.
Specifically, image-level detectors primarily focus on identifying spatial artifacts caused by forgeries. Examples include frequency anomalies~\cite{F3Net,frequency_1,frequency_2,freqblender}, identity-based discrepancies~\cite{implicit_1,implicit_2,implicit_3}, blending artifacts~\cite{X-Ray,SBI}, and localized distortions~\cite{LAA-NET,m2tr}. In contrast, video-level detectors employ a wider range of strategies. For example, some algorithms, such as ~\cite{LipForensics,eyes-blinking}, concentrate on temporal consistency, typically reflected in correlations across consecutive frames. Other video detection studies~\cite{spatiotemporal_1,spatiotemporal_2,FTCN} incorporate both temporal and spatiotemporal features to establish a more comprehensive detection framework. Another line of research~\cite{audio-visual_1,audio-visual_2} examines audio-visual consistency by analyzing the synchronization and correlation between video and audio signals to detect forgeries.
In summary, these methods rely heavily on existing artifacts in training set, limiting their generalization to unseen manipulation types. In contrast, our approach explores feature-space anomalies, enhancing the generalization capacity of our proposed detector to diverse deepfake manipulations.

\paragraph{Towards Generalization in Deepfake Detection.}
Rapid advance in deepfakes have significantly improved the quality and authenticity of generated content, posing substantial challenges to the generalization capability of detection systems. Existing detectors~\cite{xception,FF++,CNN-generated} often suffer from severe performance degradation when encountering novel forgery techniques or previously unseen manipulations. To address this issue, researchers have proposed various strategies to enhance detector generalization. 
For instance, FTCN~\cite{FTCN} improves generalization by capturing temporal coherence features in video sequences. Similarly, AltFreezing~\cite{AltFreezing} employs an alternating weight-freezing training strategy to facilitate the joint detection of spatial and temporal forgery artifacts. Another algorithm, StyleFlow~\cite{StyleFlow}, leverages the temporal dynamics of style latent vectors to identify anomalous patterns in generated videos.
More recently, data synthesis has emerged as an effective strategy for improving detector generalization. For example, FWA~\cite{FWA}  enhances training data by introducing facial warping artifacts, while Face X-ray~\cite{X-Ray} detects forgeries by identifying blending boundaries, independent of specific facial manipulation techniques. Building on this concept, SBI~\cite{SBI} synthesizes training samples by blending different views of the same pristine image to simulate generic forgery artifacts. Additionally, a series of recent studies~\cite{SeeABLE,AUNet,eccv2024fake} have further advanced this direction by employing various data synthesis techniques.
Unlike existing methods that primarily focus on image- or video-level data synthesis for deepfake detection, our approach proposes to explore unseen forgeries in feature space.

\section{Methodology}
\subsection{Preliminaries}

\paragraph{Overview.}
In this paper, we introduce \textit{FakeRadar}, to overcomes the challenges of cross-domain generalization in deepfake video detection by exploring the feature space to identify emerging manipulation patterns. Specifically, FakeRadar is built upon two key components: (i) \textbf{Forgery Outlier Probing}, which models the feature-space distributions of subclusters for training samples of both real and fake categories, and then generates cluster-conditioned outlier samples to simulate unseen forgeries, and (ii) \textbf{Outlier-Guided Tri-Training}, which jointly optimizes an adapter-inserted backbone and a triplet-class classifier to distinguish among samples from the ``Real'', ``Fake'', and ``Outlier'' classes. An overview of FakeRadar can be found in~Figure~\ref{fig:framework}.

\paragraph{Model Adaptation with a Frozen Backbone.}
The FakeRadar framework employs a pre-trained Vision Transformer (ViT-B) from CLIP~\cite{Clip} as a frozen backbone to preserve its learned semantic features. This model serves as the video encoder for input clips. To adapt it for deepfake detection, we integrate lightweight ST-Adapter layers~\cite{st-adapter} after each Transformer block, enabling parameter-efficient fine-tuning. This significantly enhances the model’s ability to capture spatial manipulations and temporal inconsistencies, which are crucial for detecting deepfake videos. Given a feature vector $x$ from the video encoder, each adapter transforms it as:
\begin{equation}
\text{ST-Adapter}(x) =x + \mathrm{ReLU}(\mathrm{Conv3D}(xW_{\text{down}}))W_{\text{up}}.
\end{equation}
\noindent
Here, $\mathrm{Conv3D}(\cdot)$ denotes a 3D convolution operation employed for capturing spatio-temporal features from videos. As well,  \(W_{\text{down}}\) and \(W_{\text{up}}\) are weight matrices that perform dimensionality reduction and expansion, respectively, and \(\mathrm{ReLU}(\cdot)\) is the activation function.
By introducing only a small number of additional parameters, these adapters allow efficient fine-tuning while retaining the original backbone’s representational strength. 
After feature extraction, each video clip is represented as a $d$-dimensional vector for subsequent analysis, with $d=768$.
 
\paragraph{Inference.} 
During inference, the model classifies input samples as either real or deepfake, with deepfakes categorized as “Fake” or “Outlier”. The “Fake” class includes fake samples with known manipulation patterns, while the “Outlier” class detects testing samples with novel forgeries that differ from established manipulations.  
By unifying these categories, the model enhances its robustness against evolving manipulation techniques, enabling accurate detection of both familiar and emerging deepfakes. This improves its generalization performance in cross-domain evaluations.

\subsection{Forgery Outlier Probing}
In this section, we introduce {\textit{Forgery Outlier Probing}} (FOP) to generate outlier samples with unseen forgeries beyond the existing real and fake videos, simulating previously unobserved manipulation patterns. Specifically, FOP first presents dynamic subcluster modeling, which performs category-specific clustering on the training samples of real and fake categories. Then, it models the distributions of these categories using a strategy based on Gaussian Mixture Models (GMMs), and follows by applying dynamic subcluster splitting and merging,  thereby achieving finer-grained subcluster partitioning. Finally, we generate cluster-conditioned outliers based on the resulting subclusters to stimulate unseen forgeries.

\paragraph{Dynamic Subcluster Modeling.}  
To capture subtle intra-class variations within real and deepfake videos, we apply category-specific clustering on the FaceForensics++ (FF++) dataset. Specifically, we treat the \emph{real} class and each of the four \emph{fake} classes (i.e. four types of deepfake videos created by different manipulation techniques~\cite{DFD, F2F, faceswap, NeuralTextures}) as five separate categories. For each category \(c\), let
\(
    \mathcal{X}^{(c)} = \{\mathbf{x}_1, \mathbf{x}_2, \dots, \mathbf{x}_N\}
\)
denote the set of feature embeddings (i.e. extracted from the video encoder). Inspired by~\cite{ddpm}, we then define a main clustering network:
\begin{equation}
    f_{\mathrm{main}}^{(c)}(\mathbf{x}_i) \mapsto \mathbf{r}_i = (r_{i,1}, r_{i,2}, \dots, r_{i,K}),
\end{equation}
which initially assumes \(K\) clusters, empirically with $K=5$, providing a ``soft'' assignment of each embedding \(\mathbf{x}_i \in \mathcal{X}^{(c)}\) to these \(K\) clusters. Here, \(r_{i,k} \in [0,1]\) and \(\sum_{k=1}^K r_{i,k} = 1\).
Each cluster can be modeled as a Gaussian distribution using a GMM, allowing us to represent the feature distribution of each category in a probabilistic manner. Specifically, for each cluster \(k\) in category \(c\), we define a Gaussian component characterized by 
    $\{\pi_k, \boldsymbol{\mu}_k, \boldsymbol{\Sigma}_k\}$,
where \(\pi_k \in (0,1)\) is the mixture weight for cluster \(k\), with \(\sum_{k=1}^K \pi_k = 1\), while \(\boldsymbol{\mu}_k\) and \(\boldsymbol{\Sigma}_k\) represent its corresponding mean and covariance matrix. These parameters enable the clustering network to estimate the likelihood of each data sample belonging to different clusters, aligning our model with a GMM-based formulation.
After that, the main clustering network needs to learn these parameters by aligning its soft assignments with GMM responsibilities:
\begin{equation}
    r_{i,k}^E = \frac{\pi_k\,\mathcal{N}\bigl(\mathbf{x}_i;\boldsymbol{\mu}_k,\boldsymbol{\Sigma}_k\bigr)}
         {\sum_{k'=1}^K \pi_{k'}\,\mathcal{N}\bigl(\mathbf{x}_i;\boldsymbol{\mu}_{k'},\boldsymbol{\Sigma}_{k'}\bigr)},
\end{equation}
where \(\mathcal{N}(\cdot;\boldsymbol{\mu}_k,\boldsymbol{\Sigma}_k)\) is the multivariate Gaussian density function. The network is trained using the KL-divergence loss:
\begin{equation}
    \mathcal{L}_{\mathrm{main}} = \sum_{i=1}^N \mathrm{KL}\bigl(\mathbf{r}_i \,\|\, \mathbf{r}_i^E\bigr),
\end{equation}
which ensures that the learned cluster assignments match the inferred GMM responsibilities.

To further refine cluster structures, we introduce a subclustering network for each cluster \(k\), which attempts to split the cluster into two subclusters if multi-modal structures are detected. A compactness loss is applied to encourage subclusters to be well-separated. This loss function penalizes subclusters that are too close to each other:
\begin{equation}
    \mathcal{L}_{\mathrm{sub}} = \sum_{k=1}^K \sum_{j=1}^2 \sum_{\mathbf{x}_i \in X_k} \tilde{r}_{i,j}\,\bigl\|\,
       \mathbf{x}_i - \tilde{\boldsymbol{\mu}}_{k,j}
    \bigr\|_2^2,
\end{equation}
where \(\tilde{r}_{i,j}\) represents the probability of \(\mathbf{x}_i\) belonging to subcluster \(j\), and \(\tilde{\boldsymbol{\mu}}_{k,j}\) is its mean.

Since the predefined cluster number \(K\) may not be optimal, we propose to dynamically adjust it through split/merge proposals, inspired by~\cite{ddpm}. Suppose a cluster \(k\) contains \(N_k\) points, \(X_k\). We \emph{split} if subclustering indicates that \(X_k\) is better viewed as two subgroups, \(\{X_{k,1},X_{k,2}\}\). The Hastings ratio is given by
\begin{equation}
    H_s = \frac{
        \alpha \,\Gamma\bigl(N_{k,1}\bigr)\,f_x\bigl(X_{k,1}\bigr)\,
        \Gamma\bigl(N_{k,2}\bigr)\,f_x\bigl(X_{k,2}\bigr)
    }{
        \Gamma\bigl(N_k\bigr)\,f_x\bigl(X_k\bigr)
    },
\end{equation}
where \(f_x(\cdot)\) is the marginal likelihood under a weak NIW prior, \(\alpha\) is the Dirichlet-process concentration parameter, and \(\Gamma(\cdot)\) is the Gamma function. 

If \(H_s\) exceeds a threshold, cluster \(k\) is split into two new clusters. The parameters for these new clusters are initialized from \(\tilde{\boldsymbol{\mu}}_{k,1}, \tilde{\boldsymbol{\Sigma}}_{k,1}\) and \(\tilde{\boldsymbol{\mu}}_{k,2}, \tilde{\boldsymbol{\Sigma}}_{k,2}\), respectively. Merging is similarly proposed if two clusters significantly overlap, with acceptance ratio \(H_m=1/H_s\). Repeating these steps yields an effective number of clusters \(K^{(c)}\), plus subclusters for each category \(c\). 

After these dynamic adjustments, we obtain the final subcluster set \( \hat{\mathcal{C}} = \{\hat{C}_1, \hat{C}_2, ..., \hat{C}_{K^{(c)}}\} \), where \(K^{(c)}\) is the refined number of clusters. Each adjusted subcluster \(\hat{C}_k\) is characterized by its updated mean and covariance:
\begin{equation}
\hat{\boldsymbol{\mu}}_k = \frac{1}{|\hat{C}_k|} \sum_{\mathbf{x}_i \in \hat{C}_k} \mathbf{x}_i, 
\end{equation}
\begin{equation}
\hat{\boldsymbol{\Sigma}} = \frac{1}{|\hat{C}_k|} \sum_{\mathbf{x}_i \in \hat{C}_k} (\mathbf{x}_i - \hat{\boldsymbol{\mu}}_k) (\mathbf{x}_i - \hat{\boldsymbol{\mu}}_k)^\top.
\end{equation}
We use online estimation for efficient training by maintaining a cluster-conditional queue with \( |\hat{C}_k| \) sample instances from each subcluster. In each iteration, we enqueue the embeddings of objects to their corresponding cluster-conditional queues and dequeue the same number of object embeddings.

\begin{table*}[h]
  \centering
  \scriptsize
  \label{tab:results}
      \begin{tabular}{@{}lccccccc@{}}
        \toprule
        \multirow{2}{*}{Method} & \multirow{2}{*}{Backbone} & \multirow{2}{*}{Input Type} & \multicolumn{5}{c}{Testing Set AUC (\%)} \\
        \cmidrule(lr){4-8}
        & & & FF++ & CDFv2 & DFDCP & DFDC & DFD \\
        \midrule
        Xception ~\cite{xception} & Xception & Frame & 99.3 & 73.7 & - & 70.9 & - \\
        Face X-Ray ~\cite{X-Ray} & HRNet & Frame & 97.8 & 79.5 & - & 65.5 & 95.4 \\
        PCL+I2G~\cite{PCL+I2G} & Resnet & Frame & 99.8 & 90.0 & 74.4 & 67.5 & - \\
        SLADD ~\cite{SLADD} & Xception & Frame & 98.4 & 79.7 & 76.0 & - & - \\
        UIA-ViT ~\cite{UIA-VIT} & Vision Transformer & Frame & 99.3 & 82.4 & 75.8 & - & 94.7 \\
        UCF~\cite{UCF} & Xception & Frame & - & 82.4 & - & 80.5 & 94.5 \\
        SeeABLE ~\cite{SeeABLE} & EfficientNet & Frame & - & 87.3 & 86.3 & 75.9 & - \\
        LSDA ~\cite{LSDA} & EfficientNet & Frame & - & 91.1 & 81.2 & 77.0 & 95.6 \\
        \midrule
        FTCN ~\cite{FTCN} & 3D ResNet & Video & 99.8 & 86.9 & - & 74.0 & - \\
        RealForensics ~\cite{RealForensics} & 3D ResNet & Video & 99.8 & 86.9 & - & 75.9 & - \\
        LTTD ~\cite{LTTD} & Vision Transformer & Video & 99.5 & 89.3 & - & 80.4 & - \\
        AltFreezing ~\cite{AltFreezing} & 3D ResNet & Video & 99.7 & 89.5 & - & - & 93.7 \\
        TALL ~\cite{TALL} & Swin Transformer & Video & 99.9 & 90.8 & - & 76.8 & - \\
        StyleFlow \cite{StyleFlow} & 3D ResNet & Video & 99.1 & 89.0 & - & - & 96.1 \\
        NACO ~\cite{NACO} & Vision Transformer & Video & 99.8 & 89.5 & - & 76.7 & - \\
        FakeRadar (Ours) & Vision Transformer & Video & 99.1 & \textbf{91.7} & \textbf{88.5} & \textbf{84.1} & \textbf{96.2} \\
        \bottomrule
      \end{tabular}
  \vspace{-5pt}
  \caption{
  Generalization comparisons of FakeRadar and state-of-the-art algorithms on \textbf{cross-dataset evaluation}, using video-level AUC (\%) as the metric.
  }
  \label{tab:cross_datasets}
\end{table*}

\paragraph{Cluster-Conditional Outlier Generation.}  
After deriving finer-grained cluster partitions for each category, we generate feature-space outliers to probe potential unknown forgeries. Motivated by~\cite{vos}, in order to ensure that the sampled outliers lie near the cluster boundary, we maintain a sufficiently small value \(\epsilon\) for outlier sampling from the \(\epsilon\)-likelihood region of the estimated cluster-conditional distribution. Therefore, for a subcluster from \(\hat{\mathcal{C}}\) indexed by \(k\), the outlier samples it generates can be formulated as follows:
\begin{equation}
    \mathcal{V}_k = \bigg\{ \mathbf{v}_k \;\bigg|\; \frac{\exp\Big(-\frac{1}{2}(\mathbf{v}_k - \hat{\boldsymbol{\mu}}_k)^\top \hat{\boldsymbol{\Sigma}}^{-1} (\mathbf{v}_k - \hat{\boldsymbol{\mu}}_k) \Big)}
    {(2\pi)^{d/2} |\hat{\boldsymbol{\Sigma}}|^{1/2}} < \epsilon \bigg\},
\end{equation}
where \(v_k \sim \mathcal{N}(\hat{\bm\mu}_k, \hat{{\Sigma}}_k)\) represents the virtual outliers sampled from the estimated Gaussian distribution of subcluster \(k\), with $d$ denoting its embedding dimensionality.  It is worth noting that \textit{during model optimization, all outliers generated here, i.e. $\overline{\mathcal{V}}=\mathcal{V}_1 \cup \cdots \cup \mathcal{V}_k \cup \cdots \mathcal{V}_{|\hat{\mathcal{C}}|}$, are classified into the ``Outlier'' class, separate from the real and fake classes}.

\subsection{Outlier-Guided Tri-Training}
To enable better performance of FakeRadar for tasks of deepfake video detection, we design a strategy called \textit{Outlier-Guided Tri-Training} for model optimization jointly combining proposed {outlier-driven contrastive loss} and {outlier-conditioned cross-entropy loss}. 
With this approach, a triplet-class classifier (integrated with the pre-trained backbone and adapters) is trained to explicitly distinguish among three categories: ``Real'', ``Fake'', and ``Outlier''. This design not only improves the model’s ability to detect known forgeries but also sensitizes it to emerging manipulation patterns.

\paragraph{Outlier-Driven Contrastive Loss.}
In this work, we propose an outlier-driven contrastive loss to enhance the model's ability to delineate class boundaries within the feature space, thereby improving feature discriminability. This loss function is designed to maximize the similarity between samples and their respective subcluster centers while minimizing their similarity to other subcluster centers. Additionally, it treats generated outliers as negative examples, enforcing low similarity between samples and outliers to strengthen boundary discrimination.

To this end, the loss is formulated using an InfoNCE-like contrastive loss given an input sample $i$:
\begin{equation}
\mathcal{L}_{\text{con}}
= -\log \frac{\exp\!\Bigl(\tfrac{s(h_i,\mu_{k_i^+})}{\tau}\Bigr)}
           {\sum_{k^-} \exp\!\Bigl(\tfrac{s(h_i,\mu_{k^-})}{\tau}\Bigr)
            + \sum_{v_j \in \overline{\mathcal{V}}} \exp\!\Bigl(\tfrac{s(h_i,v_j)}{\tau}\Bigr)},
\end{equation}
where $h_i$ denotes the feature representation of sample $i$, while $\mathcal{V}$ represents the outlier cache, providing additional hard negative samples to improve the model’s ability to distinguish boundary samples. Additionally, 
\( k^- \) represents all subclusters except the one to which sample \( i \) belongs, with its corresponding \( u_{k^-} \) denoting the centers of these subclusters, used as negative samples.
The function $s(\cdot, \cdot)$ denotes cosine similarity, which quantifies the similarity between samples, and $\tau$ is a temperature hyperparameter that controls the smoothness of similarity scaling.

\paragraph{Outlier-Conditioned Cross-Entropy Loss.}
While the proposed outlier-driven contrastive loss enforces stronger feature-space separation, we still require a concrete classification objective to handle a three-category task: real, fake, and outlier. In particular, the model must not only determine whether a video is fake or real but also assign a separate label for those samples situated near decision boundaries (i.e., outliers). This triplet-class cross-entropy loss is crucial, as it enables the model to explicitly differentiate uncertain boundary samples and label them as outliers, thereby avoiding confusion.

Concretely, the triplet-class cross-entropy loss can be defined as follows:
\begin{equation}
\mathcal{L}_{\text{cls}}
= - \sum_{c \in \{\text{Real},\,\text{Fake},\,\text{Outlier}\}}
  y_c \,\log\,p_c,
\end{equation}
where $y_c$ is the ground-truth label for samples belonging to class $c$, and $p_c$ is the model-predicted probability of samples from that class, with $c=\text{``Outlier''}$ denoting the outlier label. This loss compels the model to accurately partition each category in the feature space, especially for samples near class boundaries.

\paragraph{Total Loss.}
Finally, we combine both losses into the total objective for model optimization:
\begin{equation}
\mathcal{L}_{\text{total}}
= \mathcal{L}_{\text{con}} + \lambda\,\mathcal{L}_{\text{cls}},
\end{equation}
where $\lambda=0.5$ is a hyperparameter balancing the relative importance of classification and contrastive goals. In practice, each mini-batch comprises 16 training samples (FF++), and 16 outlier samples during model training.

\section{Experiments}
\label{Experiments}

\begin{figure*}[htbp]
    \centering
    \begin{subfigure}[b]{0.32\textwidth}
        \centering
        \includegraphics[width=\textwidth]{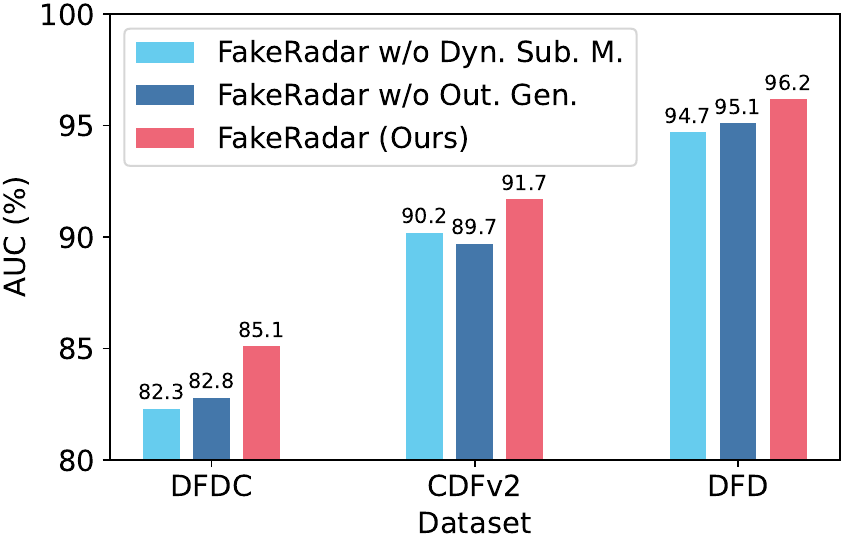}
        \vspace{2mm}
        \centerline{(a)}
    \end{subfigure}
    \hfill
    \begin{subfigure}[b]{0.32\textwidth}
        \centering
        \includegraphics[width=\textwidth]{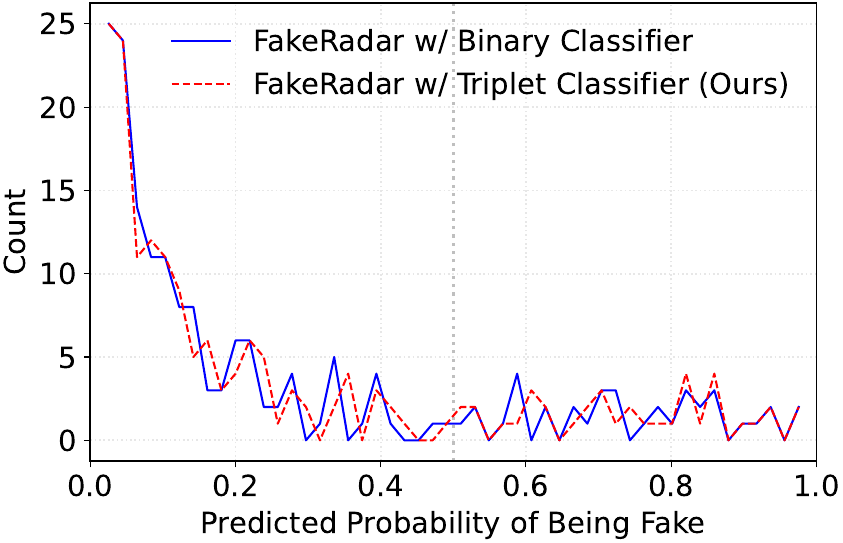}
        \vspace{2mm}
        \centerline{(b)}
    \end{subfigure}
    \hfill
    \begin{subfigure}[b]{0.32\textwidth}
        \centering
        \includegraphics[width=\textwidth]{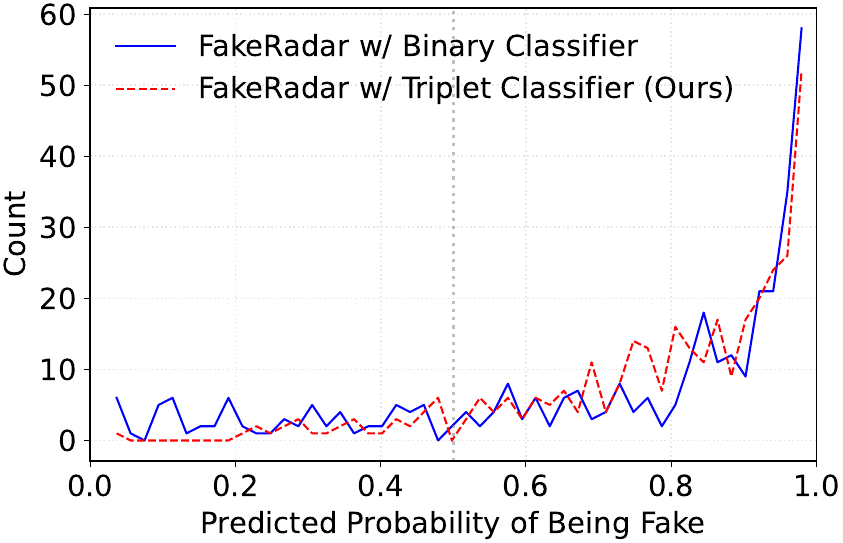}
        \vspace{2mm}
        \centerline{(c)}
    \end{subfigure}
    \vspace{-15pt}
    \caption{
    Evaluation of (a) Forgery Outlier Probing and (b)/(c) Outlier-Guided Tri-Training by FakeRadar.
    (a) ``FakeRadar w/o Dyn. Sub. M.'' and ``FakeRadar w/o Out. Gen.'' indicate that dynamic subcluster modeling and cluster-conditional outlier generation have been removed. The model is trained on FF++(HQ) and evaluated on DFDC~\cite{DFDC}, CDFv2~\cite{CDFv2}, and DFD~\cite{DFD}. (b) and (c) show, respectively, the count of real and fake test samples classified as ``fake'' across predicted probability intervals. The blue solid line denotes the binary classifier's distribution (distinguishing only real vs. fake), while the red dashed line denotes that of our proposed triplet classifier, which treats both fake and outlier samples as the new ``fake''. In this evaluation, the model is trained on FF++(HQ) and tested on CDFv2~\cite{CDFv2}.
    }
    \label{fig:combined}
\end{figure*}

\begin{table}[t]
  \centering
  \scriptsize
  \begin{tabular}{@{}ll|cccc|c@{}}
    \toprule
    Training Set & Method & DF & F2F & FS & NT & avg. \\
    \midrule
    \multirow{5}{*}{DF} 
    & DCL~\cite{DCL}  & \textbf{99.98} & 77.13 & 61.01 & 75.01 & 78.28 \\
    & UIA-ViT~\cite{UIA-VIT}  & 99.29 & 74.44 & 53.89 & 70.92 & 74.64 \\
    & TALL~\cite{TALL}  & 99.35 & 69.33 & 54.38 & 74.74 & 74.45 \\
    & AltFreezing~\cite{AltFreezing}  & 98.80 & 69.33 & 72.07 & \textbf{77.41} & 79.40 \\
    & FakeRadar (Ours) & 99.95 & \textbf{86.47} & \textbf{88.12} & 66.36 &\textbf{ 82.23} \\
    \midrule
    \multirow{5}{*}{F2F}
    & DCL~\cite{DCL}  & 91.91 & \textbf{99.21} & 59.58 & 66.67 & 79.34 \\
    & UIA-ViT~\cite{UIA-VIT}  & 83.39 & 98.32 & 68.37 & 67.17 & 79.31 \\
    & TALL~\cite{TALL}  & 85.52 & 98.67 & 62.15 & 70.24 & 79.15 \\
    & AltFreezing~\cite{AltFreezing}  & 81.12 & 98.80 & 62.97 & \textbf{70.26} & 78.29 \\
    & FakeRadar (Ours) & \textbf{96.94} & 99.09 & \textbf{82.33} & 68.46 & \textbf{86.75} \\
    \midrule
    \multirow{5}{*}{FS}
    & DCL~\cite{DCL}  & 74.80 & 69.75 & \textbf{99.90} & 52.60 & 74.26 \\
    & UIA-ViT~\cite{UIA-VIT}  & 81.02 & 66.30 & 99.04 & 49.26 & 73.91 \\
    & TALL~\cite{TALL}  & 77.83 & 67.30 & 98.77 & 52.57 & 74.12 \\
    & AltFreezing~\cite{AltFreezing}  & 80.67 & 65.94 & 99.78 & \textbf{57.23} & 75.90 \\
    & FakeRadar (Ours) & \textbf{96.39} & \textbf{82.06} & 99.83 & 52.64 & \textbf{82.73} \\
    \midrule
    \multirow{5}{*}{NT}
    & DCL~\cite{DCL}  & 91.23 & 52.13 & \textbf{79.31} & \textbf{98.97} & 80.41 \\
    & UIA-ViT~\cite{UIA-VIT}  & 79.37 & 67.89 & 45.94 & 94.59 & 71.95 \\
    & TALL~\cite{TALL}  & 84.02 & 72.85 & 51.66 & 95.16 & 75.92 \\
    & AltFreezing~\cite{AltFreezing}  & 88.81 & 76.78 & 57.92 & 95.71 & 79.80 \\
    & FakeRadar (Ours) & \textbf{96.71} & \textbf{83.34} & 67.50 & 98.41 & \textbf{86.49} \\
    \bottomrule
  \end{tabular}
  \vspace{-5pt}
  \caption{
    Generalization comparisons of FakeRadar and state-of-the-art algorithms on \textbf{cross-manipulation evaluation}, using video-level AUC (\%) as the metric.
  }
  \label{tab:cross_manipulation}
\end{table}

\subsection{Experiment Setups}
\paragraph{Datasets.} 
Following established experimental practices~\cite{AltFreezing,StyleFlow,TALL,LSDA}, this study primarily utilizes the high-quality (HQ) version of the FaceForensics++ dataset~\cite{FF++}, denoted as FF++(HQ), for FakeRadar's evaluation. This dataset consists of 1000 pristine videos and 4000 deepfakes videos generated using four different manipulation techniques: DFD~\cite{DFD}, F2F~\cite{F2F}, FS~\cite{faceswap}, and NT~\cite{NeuralTextures}. The dataset is split into training, validation, and testing sets according to the official protocol.
This study primarily focuses on evaluating the cross-domain performance of our method. To this end, we incorporate four additional datasets for evaluation, including CDFv2~\cite{CDFv2}, DFDCP~\cite{DFDCP}, DFDC~\cite{DFDC}, and DFD~\cite{DFD}.

\paragraph{Implementation Details.}
To ensure fair comparisons, we maintain most implementation details consistent with existing methods~\cite{AltFreezing,StyleFlow,NACO}. However, our FakeRadar framework employs ViT-B/16 from CLIP~\cite{Clip} as the backbone, into which we insert an ST-Adapter~\cite{st-adapter} with a 384-dimensional bottleneck for parameter-efficient fine-tuning. Additionally, the classification head of our triplet-class classifier consists of a single linear layer with softmax activation. During model training, we sample four temporal clips from each video, each containing 12 consecutive frames. We use Adam as the optimizer and apply a cosine learning rate scheduler with an initial learning rate of 1×$10^{-4}$. The model is trained for 60 epochs with a batch size of 16 and a weight decay of 5×$10^{-4}$.  
Prior to model optimization, we perform data preprocessing by detecting faces with RetinaFace~\cite{RetinaFace} and then standardize all face images to 224×224 pixels.

\paragraph{Evaluation Metrics.} 
In this work, consistent with previous studies~\cite{AltFreezing,TALL,StyleFlow} employing video-level inputs, we use the area under the ROC curve (AUC) as the performance metric during evaluations. For methods that use images as model inputs, e.g.~\cite{LSDA,SeeABLE}, their video-level results are derived by averaging the predictions across all video frames.

\subsection{Comparisons with the State-of-the-Arts}
In this study, we focus on evaluating the capabilities of cross-domain generalization for our detector to validate its effectiveness in detecting deepfake videos manipulated through various techniques. To achieve this, we employ two strategies for evaluation: cross-dataset evaluation and cross-manipulation evaluation, both of which demonstrate that our results predominantly surpass existing state-of-the-art (SoTA) algorithms for deepfake video detection, as illustrated by Table~\ref{tab:cross_datasets} and Table~\ref{tab:cross_manipulation}.

\paragraph{Cross-Dataset Evaluation.} In this situation, we train our FakeRadar on the FF++(HQ) and test its performance on multiple benchmark datasets, including CDFv2~\cite{CDFv2}, DFD~\cite{DFD}, DFDC~\cite{DFDC}, and DFDCP~\cite{DFDCP}. As reported in Table~\ref{tab:cross_datasets}, FakeRadar demonstrates its superior generalization capability in cross-dataset evaluation, achieving the best performance across all cross-dataset cases while maintaining comparable performance on the FF++ test dataset against SoTA algorithms. Specifically, our algorithm significantly improves over other methods with images and videos as inputs, particularly outperforming the current best methods, UCF~\cite{UCF} and LTTD~\cite{LTTD}, by 3.6\% and 3.7\% in the AUC metric, respectively, on the dataset DFDC~\cite{DFDC}. These results robustly showcase the proposed method's enhanced ability of generalization  to handle unseen manipulation techniques in cross-dataset evaluation scenarios.

\paragraph{Cross-Manipulation Evaluation.} To further assess our model’s generalization across different manipulation techniques, we conduct cross-manipulation evaluation on the dataset FF++(HQ). In this case, we train the model using only real samples from FF++(HQ) and one manipulation type of deepfake samples, then test its performance on the remaining three manipulation types within FF++(HQ). As shown in Table~\ref{tab:cross_manipulation}, the proposed FakeRadar achieves the best average performance in all testing cases, particularly excelling in scenarios trained with F2F as the manipulation type, surpassing the second-best performing method, DCL~\cite{DCL}, by 7.41\% in AUC. In evaluations focused on a single manipulation type, our method also achieves the best performance in most cases and the second-best in a few. This indicates that our approach, in cross-manipulation evaluation, demonstrates a performance advantage in cross-domain generalization compared to current SoTA algorithms of detecting deepfake videos.

\begin{table}[t!]
\centering
\scriptsize
\resizebox{\linewidth}{!}{
\begin{tabular}{c|ccc|cccc|c}
    \toprule
    \multirow{2}{*}{M-(\textcolor{red}{\#})} & \multirow{2}{*}{FOP} & \multicolumn{2}{c|}{OGTT} & \multirow{2}{*}{CDF} & \multirow{2}{*}{DFDC} & \multirow{2}{*}{DFDCP} & \multirow{2}{*}{DFD} & \multirow{2}{*}{avg.} \\
    & & ODCL & OCCE & & & & \\
    \midrule
    1 & \checkmark & \checkmark & \checkmark & 91.6 & 84.1 & 88.5 & 96.2 & 90.1 \\
    2 & \checkmark & \checkmark &  & 89.9 & 81.2 & 85.6 & 94.9 & 87.9 \\
    3 & \checkmark & & \checkmark & 88.6 & 80.9 & 87.7 & 94.7 & 88.0 \\
    4 & & \checkmark &  & 88.8 & 80.2 & 86.7 & 94.5 & 87.6 \\
    5 & &  &  \checkmark & 88.4 & 78.4 & 85.1 & 94.3 & 86.7 \\
    6 & &  &  & 88.2 & 78.3 & 84.8 & 94.2 & 86.4 \\
    \bottomrule
\end{tabular}
}
\vspace{-2pt}
\caption{
Ablation study results of FakeRadar by removing its one or more proposed components, using video-level AUC (\%) as the metric. 
For clarity, ``FOP'', ``OGTT'', ``ODCL'', and ``OCCE'' refer to Forgery Outlier Probing, Outlier-Guided Tri-Training, Outlier-Driven Contrastive Loss, and Outlier-Conditioned Cross-Entropy Loss, respectively. 
}
\label{tab:ablation_train}
\vspace{-0.5cm}
\end{table}

\subsection{Ablation Analysis}
\begin{figure}[t] 
    \centering
        \includegraphics[width=1\columnwidth]{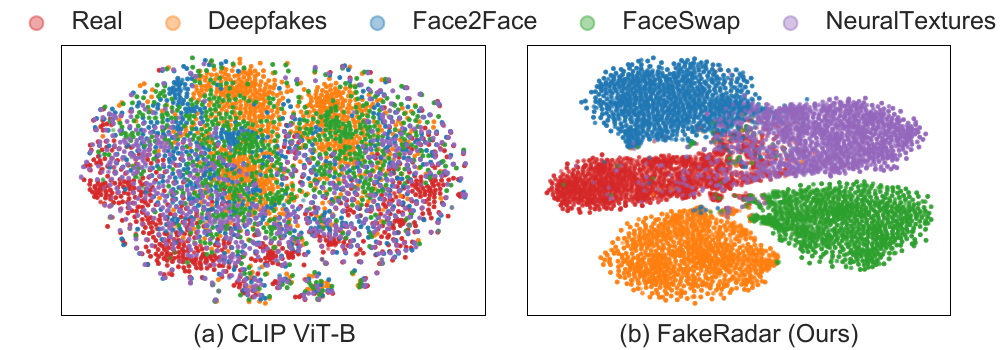} 
    \vspace{-15pt}
    \caption{Feature visualization using t-SNE. We visualize the features of (a) the pre-trained CLIP ViT-B model  and our (b) FakeRadar  here. We train both models using the training set FF++(HQ), while testing on the FF++ test set.
    } 
    \label{fig:t-SNE_FakeRadar} 
    \vspace{-15pt}
\end{figure}

We conduct an ablation study to assess the contribution of each component in FakeRadar, with results presented in Table~\ref{tab:ablation_train}, Figure~\ref{fig:combined}, and Figure~\ref{fig:t-SNE_FakeRadar}. Experiments are performed in a cross-dataset setting, where the model is trained on FF++(HQ) and tested on CDF, DFDC, DFDCP, and DFD, with average performance reported.  
As shown in Table~\ref{tab:ablation_train} (Rows M-(1) to M-(6)), removing individual or multiple components leads to a {progressive decline in AUC}, from {90.1\% to 86.4\%}, demonstrating the significance of each component. Below, we analyze the impact of FakeRadar’s key components in detail.  

\paragraph{Effectiveness of Forgery Outlier Probing.}
To evaluate the impact of {FOP}, we remove or modify its submodules: {dynamic subcluster modeling} and {cluster-conditional outlier generation}. In ``FakeRadar w/o Dyn. Sub. M.'', we disable dynamic subcluster modeling, generating outliers only from clusters of samples of real and fake categories in training set. In ``FakeRadar w/o Out. Gen.'', we replace outlier samples with Gaussian noise.  
As shown in Figure~\ref{fig:combined}(a), removing dynamic subcluster modeling reduces AUC on CDFv2, DFDC, and DFD to {90.2\%}, {82.3\%}, and {94.7\%}, respectively. While removing outlier generation has a smaller impact, performance remains below the complete FakeRadar framework. On CDFv2, {FakeRadar (91.7\%) outperforms ``FakeRadar w/o Out. Gen.'' by 2.0\%}, demonstrating that {FOP enhances generalization by generating effective outlier samples} that improve the detection of novel deepfake manipulations.  

\paragraph{Effectiveness of Outlier-Guided Tri-Training.}
Comparing Table~\ref{tab:ablation_train} (Rows M-(1) to M-(3)), removing {Outlier-Driven Contrastive Loss} or {Outlier-Conditioned Cross-Entropy Loss} leads to an {AUC drop from 90.1\% to 87.9\% and 88.9\%}, respectively, highlighting their importance. Since both losses refine the use of outlier samples, removing them would render {FOP ineffective}.  

The core of our proposed optimization strategy is the integration of a triplet classifier to handle the outlier samples generated by FOP. To assess its impact, we replace the triplet classifier, which performs three-class classification (real, fake, and outlier), with a binary classifier that only distinguishes between real and fake. We then analyze the prediction probability distributions for cross-dataset evaluation, as shown in Figure~\ref{fig:combined} (b) and (c). As illustrated, for real samples, both the binary and triplet classifiers exhibit similar prediction patterns with high confidence in the low-probability region. However, for fake samples, the triplet classifier demonstrates a clear advantage. Compared to the binary classifier, it assigns higher forgery probabilities to a greater number of fake samples, indicating that the proposed outlier-guided tri-training increases the model’s confidence in detecting unseen fake samples. This result validates the effectiveness of treating outlier samples as fake samples during inference and highlights its contribution to improving the model's generalization capability.

\paragraph{Feature Visualization.}
We use t-SNE~\cite{tsne} to visualize the features of our FakeRadar and the pre-trained CLIP ViT-B model. As shown in Figure~\ref{fig:t-SNE_FakeRadar}, the feature distribution of FakeRadar exhibits a more compact clustering structure and clearer boundaries between real and fake videos, compared to the pre-trained CLIP ViT-B model. Furthermore, FakeRadar effectively distinguishes real samples from the four types of forgery samples. This suggests that FakeRadar not only enables the model to learn discriminative features between categories but also establishes more robust decision boundaries with the aid of outlier samples. These feature distribution patterns underscore the advantages of FakeRadar in improving the model's generalization ability. The boundaries between categories are sharper, and sufficient tolerance space is provided, which is essential for addressing unseen forgery types in real-world scenarios.

\section{Conclusions}
\label{Conclusions}
In this work, we introduced FakeRadar, a novel deepfake video detection framework designed to address the limitations of existing methods in cross-domain generalization. Unlike traditional detectors that passively learn specific forgery patterns, FakeRadar proactively probes unknown forgeries by incorporating Forgery Outlier Probing to simulate diverse unseen manipulation patterns. This approach enables the model to explore a broader range of potential forgeries in feature space, significantly enhancing its adaptability to novel manipulations.
To further improve detection robustness, we proposed Outlier-Guided Tri-Training, allowing the model to distinguish between real, fake, and outlier samples. This training paradigm, introduced outlier-driven contrastive loss and outlier-conditioned cross-entropy loss, sharpens the decision boundaries and enhances the detector’s capacity to generalize across different manipulation types.
Extensive experiments on benchmark datasets confirm that FakeRadar consistently outperforms state-of-the-art detection algorithms, particularly in cross-domain evaluations where conventional detectors struggle. 

\section*{Acknowledgments}
This work was supported in part by the National Key R\&D Program of China (2024YFB3908503), in part by the Guangxi Science and Technology Project (AB25069496), and in part by the National Natural Science Foundation of China (62172120 and 62322608).

{
    \small
    \bibliographystyle{ieeenat_fullname}
    \bibliography{main}
}

\clearpage
\appendix

\twocolumn[
\begin{center}
    {\Large \bf Supplementary Material} 
    \vspace{1.5em} 
\end{center}
]

\begin{algorithm}[t!]
\caption{FakeRadar's Execution Process}
\begin{algorithmic}[1]
\item[]  \textbf{Input:} Video clips $X = \{x_1, x_2, \dots, x_n\}$ from real and fake categories
\item[]  \textbf{Initialize:} Pretrained CLIP's image encoder $\mathbf{M}$, with finetunable ST-Adapter
\For{each training iteration}
\item[] \textbf{\textcolor{blue}{// Step 1: Forgery Outlier Probing (FOP)}}
\item[]  \textbf{\textcolor{orange}{\quad\;  \textbf{Dynamic Subcluster Modeling:}}}
\For{each video clip $x_i \in X$}
    \State Extract feature $f(x_i)$ using model $\mathbf{M}$
    
    \State Partition features into subclusters using GMM
    \State Adjust subclusters through dynamical \textit{merging} and \textit{splitting} strategies
\EndFor
\item[]  \textbf{\textcolor{orange}{\quad\;  \textbf{Cluster-Conditional Outlier Generation:}}}
\For{each subcluster $C_k$}
    \State Generate outlier samples near the boundary of subcluster $C_k$
    \State Simulate unseen forgeries by generating outliers in the feature space
\EndFor
\item[] \textbf{\textcolor{blue}{// Step 2: Outlier-Guided Tri-Training}}
    \State Compute loss for each sample based on its proximity to subcluster centers
    \State Apply \textcolor{orange}{\textit{Outlier-Driven Contrastive Loss}} to separate different categories (Real, Fake, Outlier)
    \State Apply \textcolor{orange}{\textit{Outlier-Conditioned Cross-Entropy Loss}} to optimize model decision boundaries
\EndFor
\item[] \textbf{\textcolor{blue}{// Step 3: Inference}}
\For{each test sample $x_{test}$}
    \State Extract features: $f(x_{test})$
    \State Classify sample as either ``Real'', ``Fake'', or ``Outlier'' based on decision boundaries
    \If{Sample is classified as Fake or Outlier}
        \State Output: \textbf{Fake}
    \Else
        \State Output: \textbf{Real}
    \EndIf
\EndFor
\end{algorithmic}
\label{pseudocode}
\end{algorithm}

\begin{table*}[h!]
    \centering
    \small
    \begin{tabular}{lccc|ccccc}
        \toprule
        \multirow{2}{*}{Model Variant} & \multirow{2}{*}{Architecture} & \multirow{2}{*}{Input Type} & Training & \multicolumn{5}{c}{AUC (\%)} \\
        \cmidrule{5-9}
        & & & Strategy & FF++ & CDF & DFDCP & DFDC & DFD \\
        \midrule
        FakeRadar (Frozen) & ViT-B/16 & Frame & No Fine-tuning & \textcolor{gray}{55.2} & \textcolor{gray}{60.0} & \textcolor{gray}{59.0} & \textcolor{gray}{55.2} & \textcolor{gray}{57.4} \\
        FakeRadar (Supervised) & ViT-B/16+ST-Adapter & Video & Binary Classification & 98.2 & 88.2 & 84.8 & 78.3 & 94.2 \\
        FakeRadar (Proposed) & ViT-B/16+ST-Adapter & Video & FOP+OGTT  & \textbf{99.1} & \textbf{91.7} & \textbf{88.5} & \textbf{84.1} & \textbf{96.2} \\
        \bottomrule
    \end{tabular}
    \caption{Comparison of different FakeRadar variants on cross-dataset generalization, evaluated using video-level AUC (\%). The ``Frozen'' variant uses the pre-trained CLIP model without fine-tuning, while the ``Supervised'' variant integrates ST-Adapter with binary classification. The ``Proposed'' model further incorporates Forgery Outlier Probing (FOP) and Outlier-Guided Tri-Training (OGTT), leading to significant improvements across all datasets.}
    \label{tab:variant}
\end{table*}

This supplementary material provides an extended experimental exploration and in-depth analysis of our proposed model, \textit{FakeRadar}. By thoroughly examining each proposed component, we aim to gain a deeper understanding of its impact and contributions to deepfake detection. Furthermore, we present detailed experimental results and comparative analyses to validate the effectiveness of \textit{FakeRadar} under various experimental settings. \textbf{The pseudo-code of FakeRadar is provided in }\textbf{Algorithm}~\ref{pseudocode}.
The following sections outline the details of this study.

\paragraph{Effect of Model Variants on Deepfake Detection}
To systematically evaluate the impact of different components in FakeRadar (our full model, here we refer to as ``Proposed''), we design two model variants: FakeRadar (Frozen), which directly employs a pre-trained CLIP~\cite{Clip} ViT-B/16 encoder without fine-tuning; FakeRadar (Supervised), which introduces ST-Adapter~\cite{st-adapter} and binary classification, with parameter-efficient fine tuning.

As shown in Table~\ref{tab:variant}, the ``Frozen'' variant performs poorly across all datasets, achieving only 55.2\%-60.0\% AUC, demonstrating that pre-trained CLIP features alone are insufficient for deepfake detection. The ``Supervised'' variant, which incorporates ST-Adapter and binary classification, significantly improves performance, achieving 88.2\% on CDFv2~\cite{CDFv2} and 94.2\% on DFD~\cite{DFD}. However, its generalization remains limited when tested on unseen datasets.

The ``Proposed'' FakeRadar model, trained with Forgery Outlier Probing and Outlier-Guided Tri-Training, consistently outperforms both baselines. It surpasses the ``Supervised'' variant by 3.5\% AUC on CDFv2, 3.7\% on DFDCP~\cite{DFDCP}, and 5.8\% on DFDC~\cite{DFDC}, highlighting that its improved performance is attributed to the specialized training strategies rather than solely relying on a strong ViT-B backbone.

\begin{figure}[h]
    \centering
        \includegraphics[width=\linewidth]{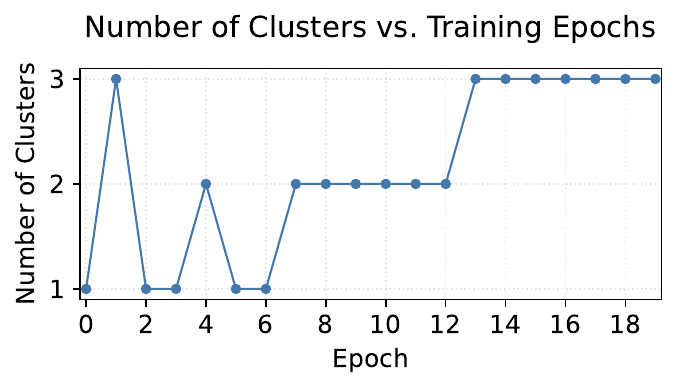}
        \label{fig:a1}
    \caption{Evolution of the number of clusters during training in Forgery Outlier Probing (FOP).  The experiment is conducted training on the FF++(HQ) training set while evaluating on DFDC. 
    }
    \label{fig:Analysis_Dynamic_Subcluster}
\end{figure}

\paragraph{How Shocking Are the Subcluster Fluctuations in Forgery Outlier Probing!}
In constructing the module of Forgery Outlier Probing, we propose a process of dynamic subcluster modeling, which involves the splitting and merging of subclusters. Using  deepfake videos of ``DFDC'' from the training set as a representative, we analyze the evolution of subcluster numbers across training epochs. 
Figure~\ref{fig:Analysis_Dynamic_Subcluster} illustrates the changes in the number of subclusters constructed during each training epoch, beginning with the initial cluster (where each manipulation type is treated as a separate cluster, and all samples from ``DFDC'' are initially assigned to the same cluster).
As shown, we observe significant fluctuations in the number of subclusters during the early training stages (Epochs 0-8). This indicates that the model is still exploring the feature distribution of the samples and gradually developing its ability to discriminate. After approximately Epoch 10, the number of subclusters stabilizes at three, suggesting that as training progresses, the model's capacity to discriminate the forgeries of clusters improves.

\begin{figure}[t]
    \centering
    \includegraphics[width=\linewidth]{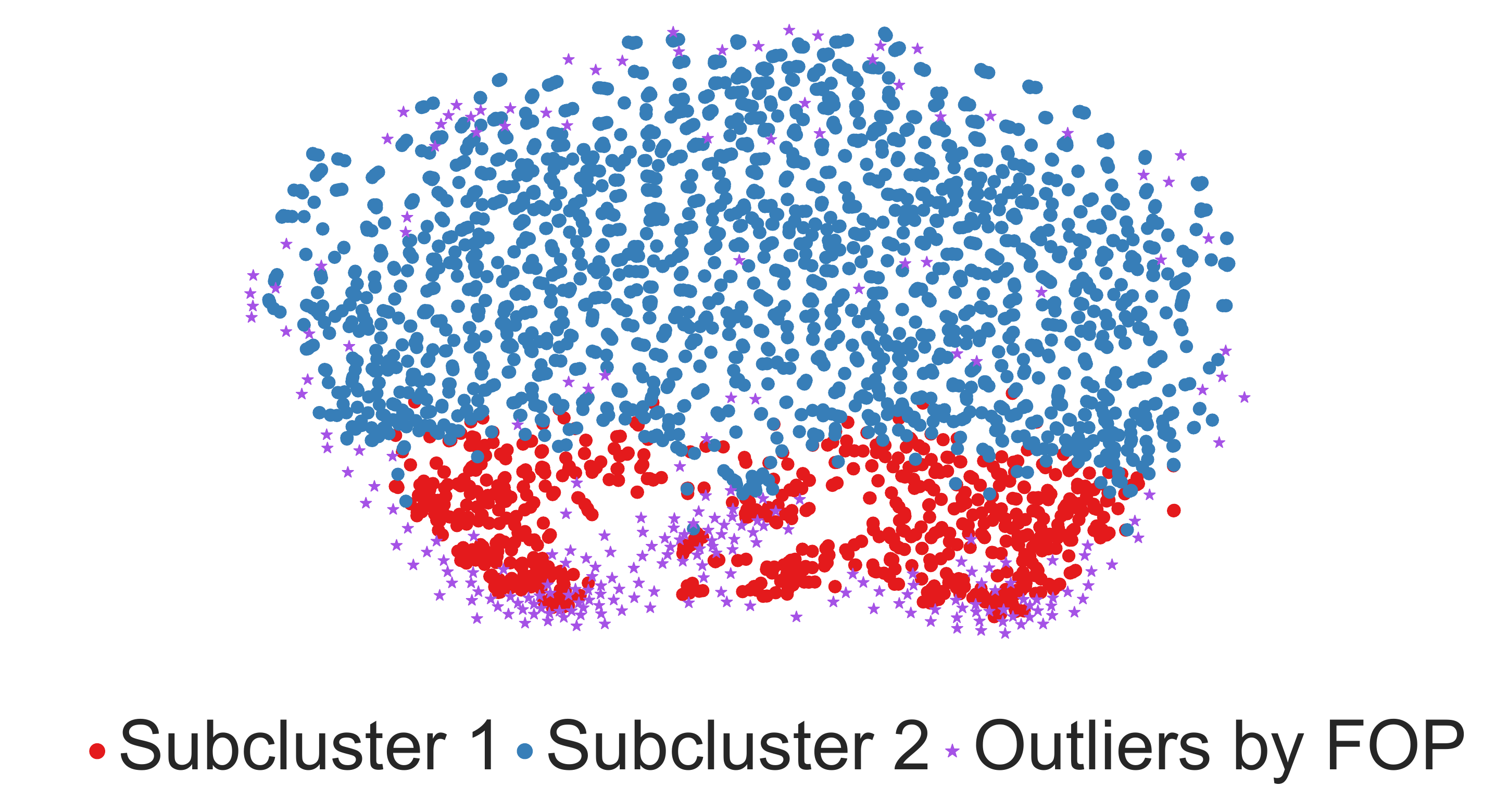}
         \vspace{-0.75cm}
    \caption{{
    t-SNE~\cite{tsne} visualization of virtual feature-space outliers generated by our Cluster-Conditional Outlier Generation. The known subclusters (``{Subcluster 1}'' and ``{Subcluster 2}'') are derived from NeuralTextures and encoded by CLIP's visual encoder~\cite{Clip}. 
    }}
\label{fig:tsne_neuraltextures}

\end{figure}

\paragraph{Visualization of Cluster-Conditional Outlier Generation.}
In this section, we visualize the virtual feature-space outliers synthesized by our Cluster-Conditional Outlier Generation approach using t-SNE~\cite{tsne} (as shown in Figure~\ref{fig:tsne_neuraltextures}). The visualization involves two known subclusters, labeled as ``Subcluster 1'' and ``Subcluster 2'', extracted from NeuralTextures and encoded using CLIP's visual encoder~\cite{Clip}. Note that, as discussed in~\cite{vos}, these synthesized outliers cannot be visualized in pixel space, as they are directly generated within a lower-dimensional feature space. As shown in Figure~\ref{fig:tsne_neuraltextures}, these virtual outliers clearly reside near the boundaries of the known subclusters, demonstrating that synthesizing unseen forgeries helps the model capture novel forgery traces beyond those of existing real data and known manipulations. This strategy thus significantly enhances FakeRadar's generalization ability in detecting previously unseen deepfake videos.

\begin{figure}[h] 
    \centering 
    \resizebox{\linewidth}{!}{
        \includegraphics[width=0.75\columnwidth]{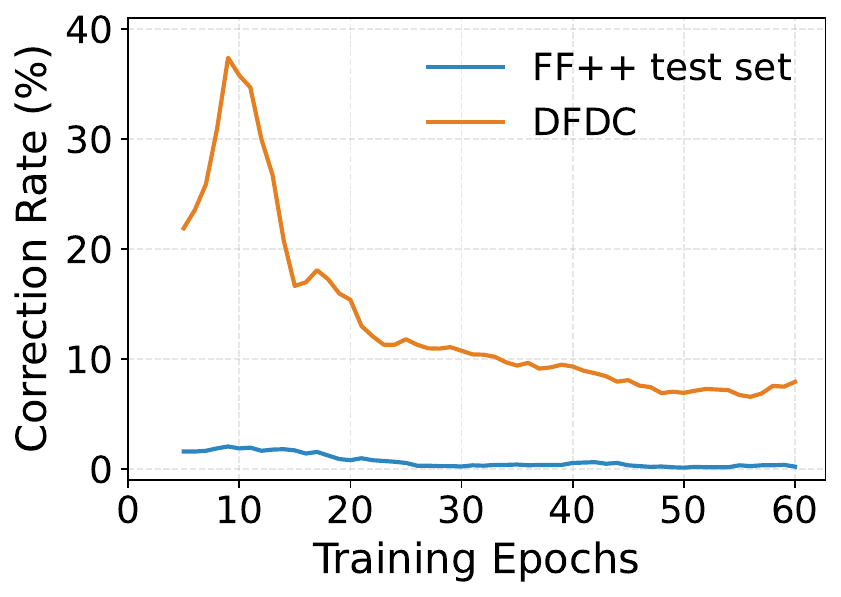} 
    }
    \caption{
    Correction rate of misclassified ``Fake'' samples over training epochs using the ``Fake'' + ``Outlier'' strategy.  The curves correspond to the correction rates on the FF++ test set~\cite{FF++} (with the same manipulation patterns as training data) and the DFDC~\cite{DFDC} dataset (with different forgery traces from those deepfakes of training data). As training progresses, the correction rates for both datasets initially increase and then decrease, indicating that the model improves in distinguishing deepfake samples over time.
    Evaluation of Outlier-Conditioned Cross-Entropy Loss on deepfake sample misclassification over training epochs. 
    The model is trained on the FF++(HQ) training set, while evaluating on the FF++ test set and DFDC.
    } 
    \label{fig:OCCEL_ablation} 
    \vspace{-10pt}
\end{figure}

\paragraph{Correcting Misclassifications in Deepfake Detection with FakeRadar.}
To validate the overall effectiveness of our proposed FakeRadar, we evaluate the impact of the core components in detection performance of deepfake videos. In our approach, the classifier is designed with three categories: ``Real'', ``Fake'' and ``Outlier''. During evaluating the model, we focus on how the triplet-class classifier corrects misclassifications of deepfake samples, specifically those misclassified as ``Fake'', by reclassifying them into a combined category of new ``Fake'' (denoted as  ``Fake'' + ``Outlier''), as illustrated in Figure~\ref{fig:OCCEL_ablation}.
In the experiment, the model is trained using the FF++(HQ) training set and evaluated on both the FF++ test set (which shares the same manipulation patterns as deepfakes of the training data) and the DFDC dataset (which contains different forgeries from deepfakes in training set). Here, we calculate the proportion of misclassified ``Fake'' samples that are subsequently corrected by the strategy of ``Fake'' + ``Outlier''.

As illustrated, the results show that during training, the correction rates for both the FF++ test set and DFDC initially increase and then decrease. Specifically, around the 10-th epoch, the correction rates reach approximately 5\% and 40\%, respectively, before declining to about 3\% and 10\% at the last. These findings suggest that as the model’s performance improves, its ability to discriminate deepfake samples enhances, thereby reducing the need for corrective reclassification.

Additionally, the similarity between the manipulation type in a sample and those in training set significantly influences the outcomes of the ``Fake'' and ``Fake'' + ``Outlier'' categories. For samples with forgery types similar to those in training data, the model assigns a high confidence level through the ``Fake'' classifier, enabling the binary classifier (Real vs. Fake) to accurately classify the sample. In these cases, the corrective effect of the ``Fake'' + ``Outlier'' strategy is limited. However, for samples with forgery types differing from those in training data, the ``Fake'' classifier assigns a lower confidence level. In such cases, the ``Outlier'' component substantially enhances prediction confidence, effectively correcting misclassifications.

To sum up, our experimental results demonstrate the following: (1) synthesizing ``Outlier'' samples to simulate unseen forgeries effectively expands the model’s exploration of unknown forgery types and corrects misclassifications within the ``Fake'' category, thereby validating the effectiveness of our proposed Forgery Outlier Probing; (2) compared to standard binary cross-entropy loss, our proposed Outlier-Conditioned Cross-Entropy Loss offers superior performance by assigning a distinct category to outlier samples, which compels the model to learn a more discriminative decision boundary and prevents misclassification of outliers as real samples; and (3) while the model’s inherent discrimination ability improves during training, our approach remains effective, particularly for forgery types that differ from those in training data, where the proposed strategy yields more significant benefits.

\paragraph{Necessity of Dynamic Subcluster Modeling.}
The goal of dynamic subcluster modeling is to uncover fine-grained patterns within each forgery category. Due to variations in source videos and manipulation techniques, each forgery type typically contains multiple distinct subgroups. Treating these heterogeneous subgroups as a single cluster often obscures low-confidence samples near category boundaries. To address this, we propose the \textbf{subclustering network}, which dynamically partitions coarse clusters with high dispersion into more coherent subclusters. Consequently, our outlier generator can sample challenging outliers around cluster boundaries, significantly enhancing the model's generalization capability to unseen forgery types.
To validate the effectiveness of our proposed method, we design two model variants for ablation analysis: 
\begin{quote}
\begin{itemize}
    \item[(1)] A fixed-subcluster variant with $K=5$, effectively disabling dynamic subcluster adjustment and directly using cluster-conditional outlier generation.
    \item[(2)] A variant trained without prior knowledge of forgery subtypes, using only two labels (\textit{Real} and \textit{Fake}), where all forgery subtypes are merged into a single \textit{Fake} class.
\end{itemize}
\end{quote}
As shown in Table~\ref{tab:TABLE1}, the fixed-subcluster variant (M-(2)) achieves an average AUC of $87.4\%$, underperforming our full FakeRadar model (M-(3), $90.1\%$) by $2.7\%$. This result highlights that dynamically adapting the number of subclusters ($K$) enables FakeRadar to better capture subtle intra-category variations, thus significantly improving generalization to unseen forgery types. Moreover, removing fine-grained subtype information only slightly reduces the cross-domain accuracy from $90.1\%$ to $89.5\%$ (M-(3) $\rightarrow$ M-(1)). This indicates that FakeRadar can still effectively generalize without forgery-specific labels, although incorporating detailed subtype information provides additional performance gains.

\begin{table}[t]
  \centering
  \resizebox{0.48\textwidth}{!}{
  \begin{tabular}{c|l|cccc|c}
    \toprule
    M-(\#) & Method & CDFv2 & DFDCP & DFDC & DFD & Average \\
    \midrule
    1 & FakeRadar (no prior) & 91.6 & 88.1 & 83.1 & 95.3 & 89.5 \\
    2 & FakeRadar (fixed $K=5$) & 90.0 & 88.3 & 82.0 & 94.1 & 87.4 \\
    3 & FakeRadar (Ours) & \textbf{91.7} & \textbf{88.5} & \textbf{84.1} & \textbf{96.2} & \textbf{90.1} \\
    \bottomrule
  \end{tabular}
  }
    \caption{Ablation analysis of FakeRadar with different subcluster modeling strategies. Results reported in AUC (\%). The best results are shown in bold.}
  \label{tab:TABLE1}
\end{table}

\paragraph{Model Robustness to Unseen Perturbations.}
Following prior work~\cite{haliassos2021lips}, we train our model on the FF++ (HQ) dataset and evaluate its robustness across various unseen perturbations at different severity levels. These perturbations include saturation changes, contrast adjustments, block-wise masking, Gaussian noise, and image compression. Robustness is measured by the average drop in AUC scores, as detailed in Table~\ref{tab:TABLE3}. 
Overall, our proposed FakeRadar demonstrates strong robustness on average across different perturbations compared to other algorithms. Notably, FakeRadar significantly outperforms competing methods under Gaussian noise (average AUC drop: $-27.9\%$) and block-wise masking (average AUC drop: $-0.2\%$), highlighting its superior resilience against challenging corruptions.

\begin{table}[t]
  \centering
  \scriptsize
  \resizebox{0.48\textwidth}{!}{
  \begin{tabular}{l|ccccc|c}
    \toprule
    Method & Saturation & Contrast & Block & Noise & Blur & Average\\
    \midrule
    AltFreezing~\cite{AltFreezing} & \textbf{-0.4} & \textbf{-0.9} & -8.0 & -38.9 & \textbf{-1.5} & -9.9\\
    StyleFlow~\cite{StyleFlow} & \textbf{-0.4} & -3.8 & -7.4 & -44.6 & -2.3 & -11.7\\
    FakeRadar (Ours) & -1.3 & \textbf{-0.9} & \textbf{-0.2} & \textbf{-27.9} & -7.8 & \textbf{-7.6}\\
    \bottomrule
  \end{tabular}
  }
    \caption{Robustness evaluation of FakeRadar and comparison methods under various perturbations, measured by average AUC drop (\%). Best results are highlighted in bold.}
  \label{tab:TABLE3}
\end{table}

\end{document}